\documentclass[lettersize,journal]{IEEEtran}
\usepackage{amsmath,amsfonts}
\usepackage{array}
\usepackage[caption=false,font=normalsize,labelfont=sf,textfont=sf]{subfig}
\usepackage{textcomp}
\usepackage{stfloats}
\usepackage{url}
\usepackage{verbatim}
\usepackage{graphicx}
\usepackage{cite}
\usepackage{booktabs}
\usepackage{amssymb}
\usepackage[ruled]{algorithm2e}
\usepackage{multirow}
\usepackage{hyperref}
\hypersetup{colorlinks=true,hypertex=true,linkcolor=red,anchorcolor=red,citecolor=green}

\usepackage{color, soul}
\usepackage{xcolor}
\usepackage[numbers,sort&compress]{natbib}  
\usepackage[linewidth=1pt]{mdframed}

\newenvironment{framedfigure*}[1][]
{
  \begin{figure*}
  \begin{mdframed}[#1]
}
{
  \end{mdframed}
  \end{figure*}
}

\begin{document}


\title{STNet: Deep Audio-Visual Fusion Network for Robust Speaker Tracking}

\author{Yidi Li, \textit{Member}, \textit{IEEE}, Hong Liu\textsuperscript{\dag}, \textit{Member}, \textit{IEEE}, Bing Yang
\thanks{
This work is supported by National Natural Science Foundation of China (No. 62403345).
Yidi Li is with the College of Computer Science and Technology, Taiyuan University of Technology, Taiyuan, China, and the Key Laboratory of Machine Perception, Peking University, Shenzhen Graduate School, China (email: liyidi@tyut.edu.cn). Hong Liu is with the Key Laboratory of Machine Perception, Peking University, Shenzhen Graduate School, China (\textsuperscript{\dag}Corresponding author: hongliu@pku.edu.cn). 
Bing Yang is with Westlake University, Westlake Institute for Advanced Study, Hangzhou, China (email: bingyang@westlake.edu.cn).
}
}

\markboth{IEEE TRANSACTIONS ON MULTIMEDIA}%
{Shell \MakeLowercase{\textit{et al.}}: A Sample Article Using IEEEtran.cls for IEEE Journals}


\maketitle

\begin{abstract}
Audio-visual speaker tracking aims to determine the location of human targets in a scene using signals captured by a multi-sensor platform, whose accuracy and robustness can be improved by multi-modal fusion methods. 
Recently, several fusion methods have been proposed to model the correlation in multiple modalities. 
However, for the speaker tracking problem, the cross-modal interaction between audio and visual signals hasn't been well exploited.
To this end, we present a novel Speaker Tracking Network (STNet) with a deep audio-visual fusion model in this work. We design a visual-guided acoustic measurement method to fuse heterogeneous cues in a unified localization space, which employs visual observations via a camera model to construct the enhanced acoustic map. For feature fusion, a cross-modal attention module is adopted to jointly model multi-modal contexts and interactions. The correlated information between audio and visual features is further interacted in the fusion model. Moreover, the STNet-based tracker is applied to multi-speaker cases by a quality-aware module, which evaluates the reliability of multi-modal observations to achieve robust tracking in complex scenarios. Experiments on the AV16.3 and CAV3D datasets show that the proposed STNet-based tracker outperforms uni-modal methods and state-of-the-art audio-visual speaker trackers.

\end{abstract}

\begin{IEEEkeywords}
Audio-visual fusion, speaker tracking, audio-visual learning, cross-modal attention.
\end{IEEEkeywords}

\section{Introduction}
\IEEEPARstart{S}{peaker} tracking is a fundamental task in human-computer interaction that determines the position of the speaker in each time step by analyzing data from sensors such as microphones and cameras \cite{zhao2023audio}.
It has wide applications in intelligent surveillance \cite{Meetings}, multimedia systems \cite{Reddy_2021_CVPR}, and robot navigation \cite{Human-robot}.
In general, the basic approaches for solving the tracking problem include computer vision-based face or body tracking methods \cite{chen2021transformer, Yoon_2019_CVPR, DetectFaces} and auditory-based Sound Source Localization (SSL) methods \cite{chen2022end, diaz2020robust}.
However, it is difficult for uni-modal methods to adapt to complex dynamic environments. 
For example, visual trackers are susceptible to object occlusion and changes in illumination and appearance. 
Besides, acoustic tracking is not subject to visual interference, but the intermittent nature of speech signals, background noise, and room reverberation constrain the performance of SSL-based trackers.
To this end, audio-visual fusion methods have become an essential approach to improve the accuracy and robustness of speaker tracking systems, taking advantage of the rich information provided by multi-modal sensors.

Most current audio-visual speaker tracking algorithms are based on the Bayesian inference framework for multi-sensor data fusion, which approximates the state distribution of the target from noisy multi-modal measurements \cite{liu2018non, vermaak2001sequential, VK2016TMM,  gatica2003audio, checka2004multiple, lyd2019icip,brutti2010joint, Qian2018ICASSP}. 
Among them, Particle Filter (PF) is a representative method due to its robustness in nonlinear non-Gaussian systems and flexibility in supporting multiple types of features.
The PF framework is applicable to fuse independently generated visual and acoustic observations with a likelihood model, usually in the form of multiplicative or additive fusion \cite{gatica2003audio,brutti2010joint, checka2004multiple, lyd2019icip, Qian2018ICASSP}.
Visual features are usually modeled based on manual features such as color histograms and appearance representations of the detected objects \cite{Qian2018ICASSP, zhou2008target, ban2017exploiting, ban2019variational, talantzis2008audio, brunelli2006generative}. 
Meanwhile, acoustic signals from microphone arrays are used to extract multi-channel spectral cues to generate SSL estimates such as Time Difference of Arriva (TDOA) \cite{d2012person, zotkin2002joint, ImprovedGCF} and Direction of Arrival (DOA) \cite{kirchmaier2011dynamical, gebru2015audio, barnard2014robust}.
The DOA angle estimates are projected onto the image plane to guide the particle propagation step and reweight the visual posterior \cite{VK2015TMM, VK2016TMM}.
3D visual observations obtained from face detectors and geometric transformations are used to assist acoustic maps in computing multi-modal likelihood components \cite{qian2019TMM, Qian2022TMM}.
The visual and audio particles distributed in two spaces are fused in an improved likelihood by a two-layer PF \cite{lyd2019icip}.
In GAVT, a multi-pose face detector is used to improve the visual likelihood, and a probabilistic steered response power is designed to enhance the audio observation model \cite{sanabria2023audiovisual}.
However, these methods prefer to use the detection of a single modality to refine the observation model of another modality, which leads to a dependence on the results of the dominant modality while ignoring the effect of unreliable noisy observations.
In addition, likelihood models using standard multiplicative or additive fusion do not provide much flexibility in capturing complex relationships between different modalities and lack low-level cross-modal information interaction.

With the development of deep learning, multiple neural networks have been widely proposed for multi-modal tasks, bringing great inspiration to audio-visual fusion study \cite{Multimodal2019tpami}.
Deep learning-based audio-visual fusion methods have been proven to enhance many machine perception tasks. They are in principle able to address the challenges of data uncleanness, noise, missing, and conflicts \cite{katsaggelos2015audiovisual}.
The Convolutional Neural Network (CNN) is introduced to build an audio-visual object tracking framework, where an additive merging layer is designed to fuse audio and visual features before the standard classification network \cite{AVOT2020icra}.
A multi-channel Deep Neural Network (DNN) is proposed for sound source localization in the image scene, including a visual DNN for localizing candidates and an audio DNN for voice activity detection \cite{masuyama2020self}.
Deep learning methods considerably improve the performance of audio-visual models. However, it is commonly used for feature extraction or related auxiliary tasks.
There is a lack of an end-to-end deep audio-visual learning-based tracking framework specifically for the speaker tracking task.

In this paper, we propose a novel Speaker Tracking Network (STNet) with a deep audio-visual fusion model.
In the feature extraction step, the input representations of audio and visual modalities with different noise topologies show significant differences.
Therefore, we embed heterogeneous signals into a unified localization space and explore audio-visual representations across both domains.
The acoustic maps that are projected onto the image plane are fed into a CNN-based architecture designed for image processing.
For audio-visual feature fusion, we employ the cross-modal attention mechanism to facilitate the interaction of information from audio and visual streams.
STNet utilizes the powerful spatial feature extraction capability of CNN while modeling the correlation between audio and visual features through an attention module, extracting fine-grained local features while modeling long-range global contexts across modalities.
Further, the network is extended to multi-speaker tracking scenarios via a quality-aware module.
Within this, the quality scores of the network outputs are estimated by combining information from uni-modal and fusion features.
The quality-aware-based update and reset strategy ensures the continuity of multi-target matching and tracking.
Finally, all these components make up our STNet-based Tracker (STNT), which achieves state-of-the-art results on the AV16.3 and CAV3D datasets.

The contributions of this paper are summarized as follows:
\begin{itemize}
\item We propose a novel deep audio-visual fusion network to address the speaker tracking problem. The proposed STNet comprehensively considers the correlation and interaction among audio and visual feature embeddings via a cross-modal attention module.
\item We design a visual-guided acoustic measurement method that develops a mapping relationship between the audio and visual representations, realizing the fusion of heterogeneous cues in a unified localization space. 
\item We propose a quality-aware module to handle multi-speaker tracking via an update and reset strategy.
\item Experimental results show that the proposed tracker outperforms state-of-the-art algorithms on the commonly used benchmark dataset AV16.3 and CAV3D.
\end{itemize}
\section{Background}
In this section, we briefly review the related effective methods in two areas: audio-visual speaker tracking and audio-visual fusion. Then we introduce an acoustic measurement method that is the basis of the proposed audio-visual measurement method.

\subsection{Audio-Visual Speaker Tracking}
Using multi-modal information is an essential approach to enhance tracking performance because the complementary nature of audio and visual modalities can be used to improve the respective strengths and compensate for the individual weaknesses in tracking.
Audio measurements for audio-visual speaker tracking systems are typically derived using SSL algorithms, such as delay estimation and DOA estimation \cite{ImprovedGCF,d2012person, zotkin2002joint,kirchmaier2011dynamical, gebru2015audio, barnard2014robust}.
Meanwhile, visual cues are typically derived using either hand-crafted generative appearance features \cite{VK2016TMM, zhou2008target, ban2017exploiting, ban2019variational, brutti2010joint, talantzis2008audio, brunelli2006generative} or deep neural network-based discriminative features \cite{gebru2017audio, AVOT2020icra}.
For audio-visual speaker tracking, the state space approach based on the Bayesian framework is a representative model-driven approach due to its strength in fusing multi-modal information \cite{gatica2003audio}.
PF is the most commonly used tracking framework, which makes approximate inferences of the filtering distribution of the target and exhibits robustness in nonlinear non-Gaussian systems \cite{gatica2003audio, checka2004multiple, VK2015TMM, lyd2019icip,EUSIPCO2020, LYD-ICPR}.
The Probability Hypothesis Density (PHD) filter is proposed to overcome the problem of the unknown and varying number of speakers by propagating multiple target posteriors \cite{VK2016TMM, liu2018non}.
The Poisson Multi-Bernoulli Mixture (PMBM) filter is introduced with a phase-aware filter and a separation-before-localization method for multi-speaker tracking \cite{zhao2022PMBM, ImprovedGCF}.
The graphical model solved by variational approximations is proposed to deal with the observation missing problem due to intermittent streams \cite{gebru2017audio, ban2019variational}.
The deep learning method is widely used in multi-modal fusion studies \cite{qian2021audio, yu2021mpn, tian2018audio} but rarely introduced to audio-visual tracking due to insufficient labeled data \cite{AVOT2020icra,lyd2022AAAI}.
For audio-visual robotic speaker DOA estimation, a cross-modal attentive fusion mechanism is proposed to explore intra-modal temporal dependencies and inter-modal alignment \cite{cmaf}.
A multi-modal perceptual attention module is trained by a cross-modal self-supervised learning method \cite{lyd2022AAAI}. 
This module explicitly models the multi-modal perceptual process and is further combined into a PF-based tracking framework.
Instead of the combination of feature extraction and Bayesian filter, in this paper, we design an end-to-end speaker tracking paradigm based on deep audio-visual learning.

\subsection{Audio-Visual Fusion}
The trend from uni-modal learning to multi-modal learning is crucial for the development of machine perception.
Based on the two most significant perceptual modalities, visual and auditory, audio-visual learning aims to exploit the relationship between audio and visual modalities. 
It is introduced to many challenging tasks, such as audio-visual separation \cite{r3-overview, r3-www2013Separation, r3-www2005separation, gabbay2018seeing, Cocktail2018}, object localization \cite{arandjelovic2018objects, tian2018audio, pu2017audio, wu2021binaural}, audio-visual speech recognition \cite{afouras2018deep, ma2021conformers}, and audio and visual generation \cite{lin2021exploiting, liu2021towards, shlizerman2018audio}.
Taking advantage of deep learning, modality-specific features and joint representations are learned implicitly in increasingly advanced fusion schemes.
The binaural audio-visual network extracts and integrates features from binaural recordings and images concurrently to produce a pixel-level SSL map \cite{wu2021binaural}.
The convolution-augmented transformer proposed in \cite{ma2021conformers} extracts features directly from the raw pixels and audio waveforms and fuses them via a multi-layer perceptron.
The dual multi-modal residual network is adopted to update the separately processed audio and visual features and extract supplementary useful information \cite{tian2018audio}.
Popular learning paradigms typically extract uni-modal features respectively and then learn the joint representation at the intermediate layer or perform late fusion at the decision layer.
On the contrary, we use cross-modal guidance at the observation stage to obtain more reliable cues and focus on audio-visual interactions in the fusion network.

With its impressive performance in Natural Language Processing (NLP) and Computer Vision (CV), the transformer has recently been extended to numerous multi-modal applications as a generic perceptual model \cite{jaegle2021perceiver}.
As the core idea of the transformer, attention operation provides a natural mechanism for multi-modal signals to connect and interact.
Diverse forms of attention mechanisms are applied to various multi-modal models \cite{ma2021conformers, serdyuk2021audio, LYD-transformer, song2022multimodal, lyd2022AAAI}.
In this work, we introduce a cross-modal attention module to enhance the information interaction between audio and visual modalities. 
This fusion module assigns distinct attention weights to local features of audio and visual embeddings to derive a more comprehensive audio-visual representation.
\vspace{-0.1cm}
\begin{figure*}[ht] 
\centering
\includegraphics[width=2\columnwidth]{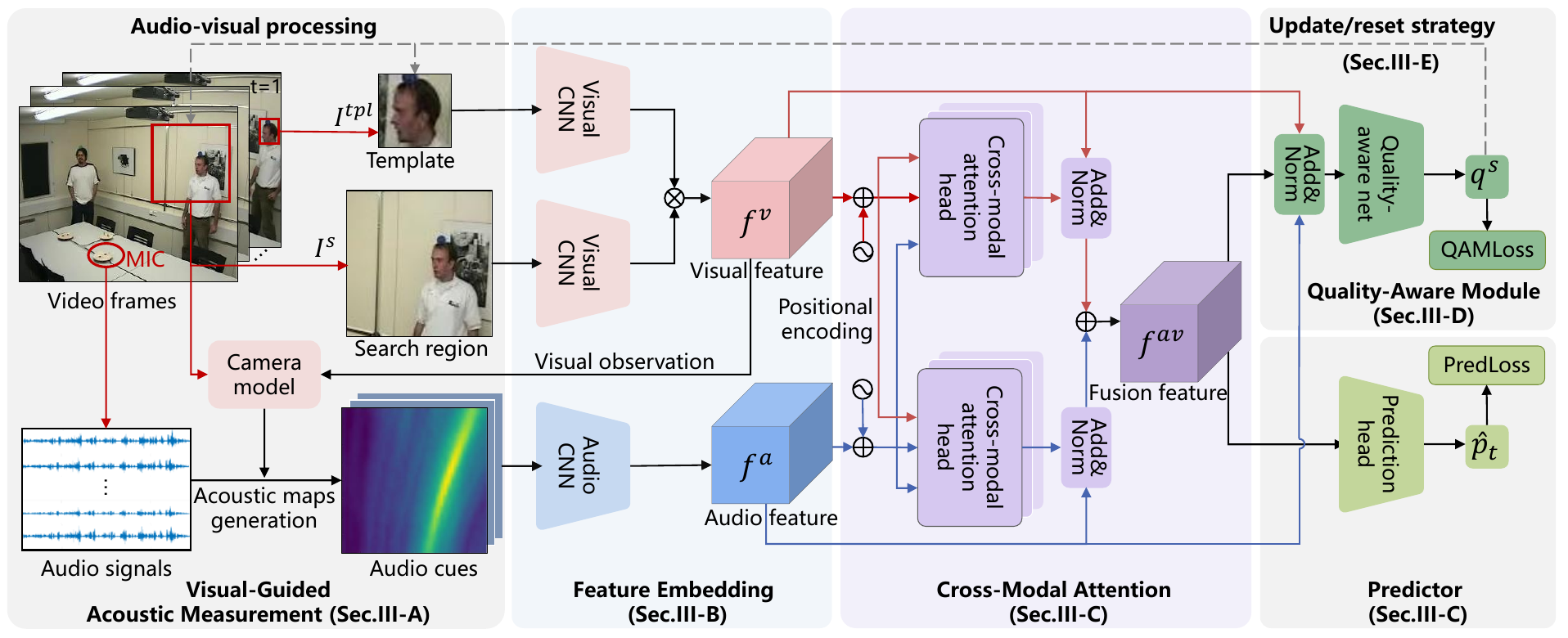} 
\caption{Network architecture of the proposed STNet. In the audio-visual processing, audio cues are derived by visual-guided acoustic measurement. Audio and visual features are extracted by the audio CNN and the Siamese-like visual CNN, and then interacted and fused in a cross-modal attention module. A quality-aware module is used to construct an update/reset strategy for multi-speaker tracking. 
The speaker position is estimated by the prediction head.
}
\label{fig-fw}
\end{figure*}

\subsection{Acoustic Measurement}
TDOA estimation is the most widely used acoustic measurement method for SSL task \cite{d2012person, zotkin2002joint, ImprovedGCF}, which infers the source position based on the estimation of the time difference between wavefronts arriving at each microphone pair and the geometry of the array.
The most prevalent approach is based on Generalized Cross-Correlation Phase Transform (GCC-PHAT), which simply discards the magnitude information and keeps only the phase information, making it more robust in noisy and reverberant situations \cite{gccphat}.
Based on coherent measurements of the cross-power spectral phase, the Global Coherence Field (GCF) feature is calculated as the average of GCC-PHATs for each microphone pair at spatial points.
The acoustic map constructed based on GCF can derive the most likely source position from the spatial peaks.
Recently, an acoustic measurement method named spatial-temporal GCF (stGCF) is proposed \cite{lyd2022AAAI}, which
utilizes a camera model to construct spatial sampling points and derive the optimal value over a time period according to movement continuity.
However, the peak distribution of stGCF is interfered by multiple synchronized sources and has high search complexity.
To this end,
we improve stGCF by using reliable visual observations to guide the sampling process to derive a more accurate acoustic map.

\section{STNet Neural Architecture}
In this work, we combine the video frames captured by a monocular camera and the audio signals collected by microphone arrays in an end-to-end framework to localize the position of the specific speaker. 
The framework of the proposed STNet is illustrated in Figure~\ref{fig-fw}, including the visual-guided acoustic measurement, feature embedding, cross-modal attention, quality-aware module, and prediction head.
First, audio cues are derived by the visual-guided acoustic measurement.
The image and speech signals are mapped from mutually independent spaces to a unified feature space through the audio-visual processing and feature embedding. 
Then, the global interaction of audio and visual features is learned in a cross-modal attention module for feature fusion.
Finally, a prediction head and a quality-aware module are used to estimate the speaker position and evaluate the task quality information.
Let $\mathbf{I}_{1:t}$ denote the image frame set, and $\mathbf{S}_{1:t}$ denote the set of windowed audio signals, a video sequence is composed of time-synchronized audio-visual sample pairs $(s_t, I_t)$, $t = 1,...,T$, where $T$ is the total number of frames. We aim to estimate the face position of the speaker at time $t$: $\hat{p}_t=(\hat{u}_t, \hat{v}_t)$, where $\hat{p}_t$ is the coordinate of the center of the target face in the image plane. In this section, we describe the components of STNet and the tracking algorithm in detail. 

\subsection{Visual-Guided Acoustic Measurement}
\label{AO}
In the audio-visual processing, we extract the stGCF-based acoustic maps \cite{lyd2022AAAI}, whose peak distribution indicates the location of the sound source. 
However, the distribution is subject to interference from ambient noise and reverberation.
Therefore, we use visual guidance to obtain an improved acoustic map in cases where reliable observations are obtained in the visual branch.
Face detection results are used to guide spatial sampling instead of the entire image frame.
Visual guidance further narrows the search area and improves SSL accuracy by excluding noise interference from other areas.

\begin{figure}[t] 
\centering
\includegraphics[width=1\columnwidth]{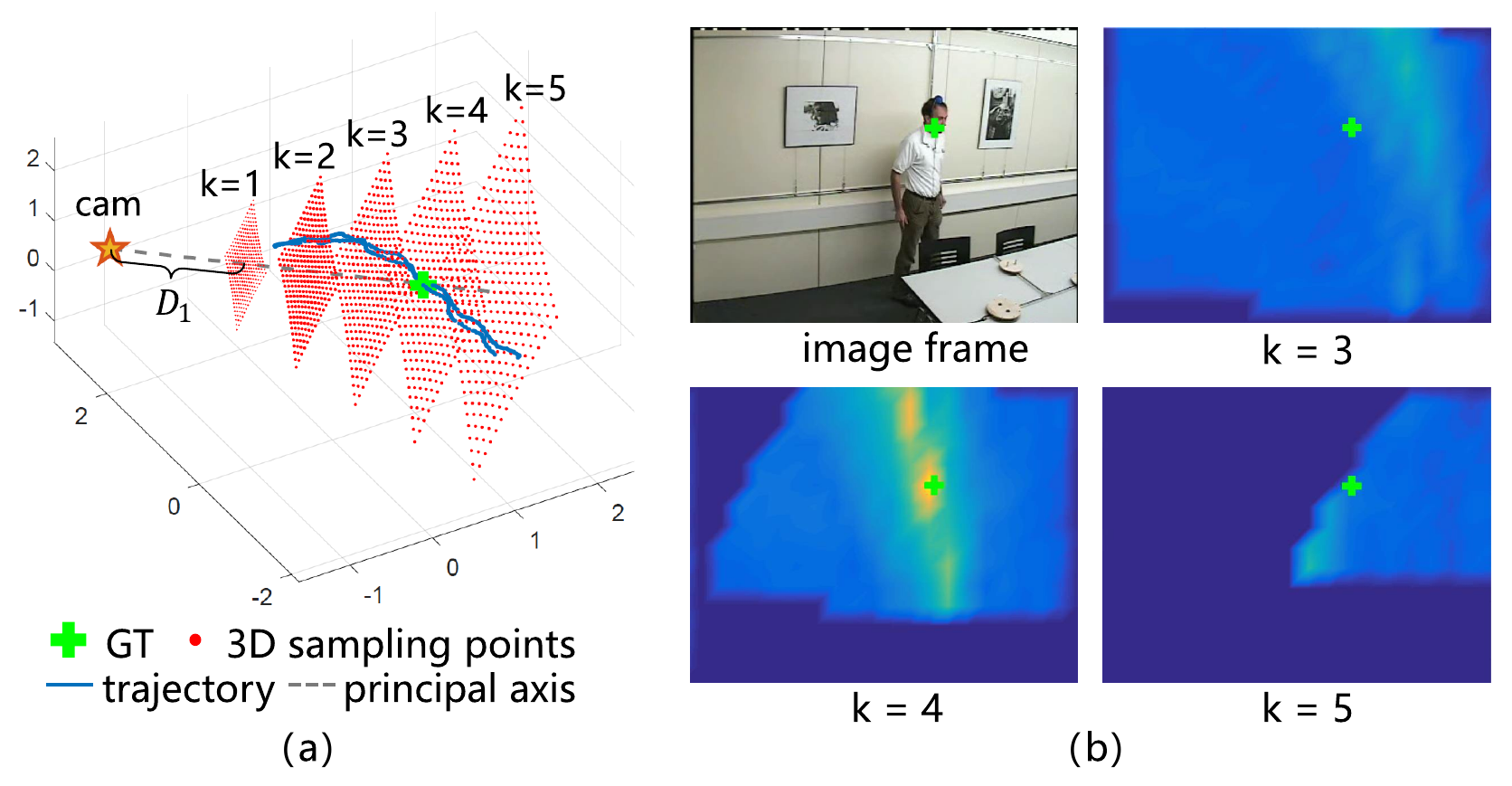} 
\vspace{-0.6cm}
\caption{(a) 3D sampling points at five depths, entire target trajectory and current ground-truth (GT).
(b) Current image frame and GCF maps at three depths. The green cross indicates the speaker's position. Yellow (blue) indicates a higher (lower) probability of source presence.}
\vspace{-0.25cm}
\label{fig-camera}
\end{figure}

Firstly, to construct the spatial grid, the 2D sample point set $\mathbf{P}^{2d}(I_t)$ is obtained by sampling on the image frame $I_t$. 
Then, the 2D points in the image plane are projected into a set of 3D points with different depths in 3D world coordinates through a pinhole camera model. Figure~\ref{fig-camera} (a) visualizes the 3D sampling points at five depths. Assume that the set of $d$ depths is denoted as $D=\{D_k, k=1,...,d\}$, and the image-to-3D projection process on depth $D_k$ is defined as:
\begin{equation}
\label{eq:projection}
\mathbf{P}^{3d}_k(I_t)=\Phi(\mathbf{P}^{2d}(I_t);D_k),
\end{equation}
where $\Phi$ is the projection operator.
Given the sampling point set $\mathbf{P}^{3d}_k(I_t)$ of image frame $I_t$ at the given depth $D_k$, the formula of GCF map is defined as:
\begin{equation}
\label{eq:GCFmap}
\mathbf{R}_{\Omega}(t,\mathbf{P}^{3d}_{k}(I_t)) = \begin{bmatrix}
\mathbf{r}(t,p_{11k}) & \dots   & \mathbf{r}(t,p_{1wk})\\
   \vdots                    & \ddots & \vdots                       \\
\mathbf{r}(t,p_{h1k}) & \dots    & \mathbf{r}(t,p_{hwk}) 
\end{bmatrix}_{h \times w},
\end{equation}
where $\mathbf{r}(t,p)$ denotes the GCF value calculated at position $p$ (see \cite{lyd2022AAAI} for more details). $k$ is the index of the depth, $h$ and $w$ are the number of sampling points in the vertical and horizontal directions. 
Considering GCF maps at $d$ depths in time interval $[t-m_1, t]$, we select $m_2$ maps with top peak values to form stGCF maps:
\begin{equation}
\label{eq:stGCF}
\mathbf{R}_{\Omega}^{st}(t,I_t) = \{\mathbf{R}_{\Omega}(t',\mathbf{P}_{k'}^{3d}(I_t))|t'\in \mathbf{T'},k'\in \mathbf{K'}\}, 
\end{equation}
where $\mathbf{T'}$ is the time index set of the $m_2$ frames, and $\mathbf{K'}$ is the corresponding depth index set.
Figure~\ref{fig-camera} shows the 3D sampling points at different depths and the resulting GCF maps at three of the depths.
In Figure~\ref{fig-camera} (b), the peaks (yellow) near the speaker's location (green cross) in acoustic maps indicate a higher probability of sound source presence. 
However, some pseudo points appear at other positions due to noise or geometric configuration of the sensors.
Therefore, we perform spatial sampling only in the face region.
When the face bounding box $O_t$ exists, the region is used to calculate the stGCF by Equation \ref{eq:stGCF}, which is denoted as $ \mathbf{R}_{\Omega}^{st}(t,O_t)$.
After deriving the corresponding time index set and depth index set, the complete GCF maps are calculated in the region of the image frame. The face-guided GCF (fgGCF) maps are calculated as:
\begin{equation}
\label{eq:fg-GCF}
\mathbf{R}_{\Omega}^{fg}(t,O_t,I_t) = \{\mathbf{R}_{\Omega}(t^o,\mathbf{P}_{k^o}^{3d}(I_t))|t^o\in \mathbf{T}^o,k^o\in \mathbf{K}^o\}, 
\end{equation}
 where $\mathbf{T}^o$ and $\mathbf{K}^o$ represent the time and depth index set to construct $ \mathbf{R}_{\Omega}^{st}(t,O_t)$. Finally, the visual-guided GCF (vgGCF) is defined as:
\begin{equation}
\label{eq:vg-GCF}
\mathbf{R}_{\Omega}^{vg}(t,I_t)=
\left\{
\begin{array}{l}
\mathbf{R}_{\Omega}^{fg}(t,O_t,I_t)\ \ \ \mathrm{if} \ O_t \neq \varnothing \\
\mathbf{R}_{\Omega}^{st}(t,I_t)\ \ \ \ \  \ \ \ \mathrm{else}.
\end{array}
\right.
\end{equation} 
The overall visual-guided acoustic measurement process is depicted in Algorithm \ref{alg:vg-GCF}.

\begin{algorithm}
  \caption{Visual-guided Acoustic Measurement}
  \label{alg:vg-GCF}
  \SetAlgoLined
  \KwIn{$t$, $I_{t}$, $D_k$, $s_{t-m_1:t}$, $O_t$, $T$, $m_1$, $m_2$}
  \KwOut{$\mathbf{R}_{\Omega}^{vg}(t,I_t)$: visual-guided acoustic maps}

  $\mathit{Algorithm 1.1}$: compute $\mathbf{R}_{\Omega}^{st}(t,I_t)$ with $\forall t \in T$\;
  
  Construct 2D sampling points $\mathbf{P}^{2d}(I_t)$ and derive 3D sampling points $\mathbf{P}_k^{3d}(I_t)=\{ p_{11k},...,p_{hwk}\}$ for $I_t$ with Eq.\ref{eq:projection}.\;
  
    \For{$\tilde{t}=t-m_1,...,t$}{
        \For{$k=1,...,d$}{
            \For{$i=1,...,h$, $j=1,...,w$}{
              $\mathbf{r}(\tilde{t},p_{ijk}) \leftarrow s_{\widetilde{t}}, p_{ijk}$, refer to \cite{lyd2022AAAI}
            }
            $\mathbf{R}_{\Omega}(\tilde{t},\mathbf{P}^{3d}_{k}(I_t)) \leftarrow \mathbf{r}(\tilde{t},p_{\cdot \cdot k})$ with Eq.\ref{eq:GCFmap}
        }
    }
    From a group of $\mathbf{R}_{\Omega}(\tilde{t},\mathbf{P}^{3d}_{k}(I_t)) $, select $m_2$ maps with top peak values to construct $\mathbf{R}_{\Omega}^{st}(t,I_t)$ and get $\mathbf{T'}$, $\mathbf{K'}$ with Eq.\ref{eq:stGCF}
    
    $\mathit{Algorithm 1.2}$ : compute $\mathbf{R}_{\Omega}^{vg}(t,I_t)$ with $\forall t \in T$\;
    
    \eIf{$O_t \neq \varnothing$}{
      calculate $\mathbf{R}_{\Omega}^{st}(t,O_t)$ with $\mathit{Algorithm 1.1}$ and get $\mathbf{T}^o$, $\mathbf{K}^o$\;
      
      $\mathbf{R}_{\Omega}^{fg}(t,O_t,I_t) \leftarrow \mathbf{T}^o, \mathbf{K}^o $ with Eq.\ref{eq:fg-GCF}\;
      
      $\mathbf{R}_{\Omega}^{vg}(t,I_t)=\mathbf{R}_{\Omega}^{fg}(t,O_t,I_t)$\;
      
      }{
      $\mathbf{R}_{\Omega}^{vg}(t,I_t)=\mathbf{R}_{\Omega}^{st}(t,I_t)$ with $\mathit{Algorithm 1.1}$\;
      }
\end{algorithm}

\begin{figure}[t] 
\centering
\includegraphics[width=1\columnwidth]{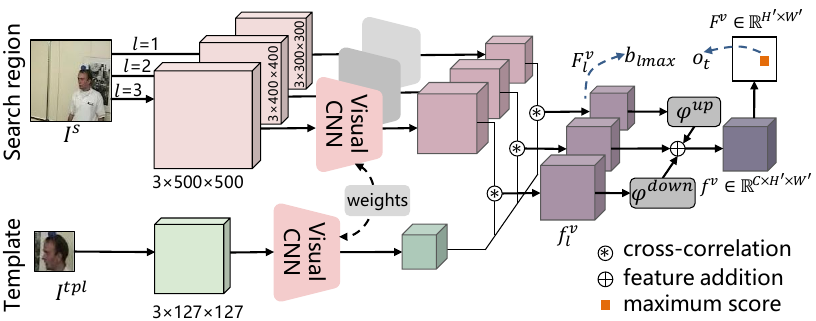} 
\caption{
The structure of the Siamese-based visual network with multi-scale fusion.
The network weights of search region and template branch are shared.}
\label{fig-vnet}
\vspace{-0.1cm}
\end{figure}

\subsection{Feature Embedding}
In this module, we use a two-stream network architecture including a visual branch and an audio branch to process image frames and audio cues, respectively.
\subsubsection{Visual network}
\label{VO}
The visual tracking problem is typically solved by similarity matching between the target template and the search region. Therefore, a Siamese-based \cite{bertinetto2016fully} network as shown in Figure~\ref{fig-vnet} is employed to capture visual observations.
A pair of image patches, the template image patch $I^{tpl}$ and the search region image patch $I^s$, is fed into a weight-sharing fully convolutional network ($\mathit{Conv}$) to extract features. Then, the cross-correlation operation is performed to obtain a response map, representing the probability of the template at each location in the search region. 
Further, we utilize a multi-scale fusion layer to handle the scale variation of targets. As shown in Figure~\ref{fig-vnet}, the multi-scale features extracted from the reshaped search images are added after up-sampling and down-sampling. 
In terms of three scales, the process of the visual network is summarized as follows:
\begin{align}
&f_l^v=\mathit{Conv^v} (I^{tpl})*\mathit{Conv^v} (I_l^{s}),l=1,2,3,\\
&f^v=\varphi^{up}(f_1^v) \oplus f_2^v \oplus \varphi^{down}(f_3^v),
\end{align}
where $l$ is the index of the scale, $*$ and $\oplus$ denote the operations of cross-correlation and feature addition. $\mathit{Conv^v}$ refers to the visual CNN similar to the five $\mathit{Conv}$ layers of Alexnet \cite{krizhevsky2012imagenet}. 
$f^v\in \mathbb{R} ^{C\times H'\times W'}$ represents the multi-scale visual features where $C=256$. $\varphi^{up}$ and $\varphi^{down}$ denote up-sampling and down-sampling operations, respectively. 
We sum the feature vector $f^v$ over the channel dimension to get the score map $F^v\in \mathbb{R} ^{H'\times W'}$.
The visual observation $O_t$ is derived as follow:
\begin{equation}
\label{eq:ot}
O_t=
\left\{
\begin{array}{l}
(o_t, b_{lmax}),\ \ \mathrm{if} \ max(F^v)\ge \theta^v   \\
\varnothing\ \ \ \ \ \ \ \ \ \ \ \ \ \,\mathrm{else},
\end{array}
\right.
\end{equation} 
where $o_t$ is the center of the face bounding box, which is derived from the position with the maximum score in the score map $F^v$. $b_{lmax}$ is the scale of the bounding box, where $lmax=argmax_{l=1,2,3}(F^v_l)$. $F^v_l$ is the score map of each feature $f^v_l$.
When the maximum value of the map $F^v$ is higher than the set threshold, the face bounding box is reliable enough to be used in visual-guided acoustic measurement.
\begin{figure}[t] 
\centering
\includegraphics[width=0.95\columnwidth]{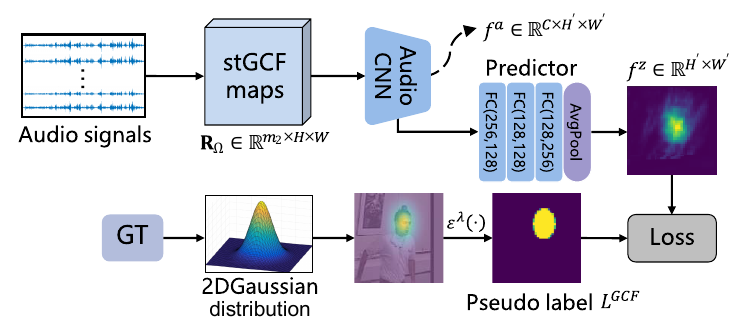} 
\vspace{-0.15cm}
\caption{
The structure of GCFNet. The bottom line shows the binary pseudo-label generated on the image frame with GT using Gaussian distribution.
}
\label{fig-GCFnet}
\vspace{-0.1cm}
\end{figure}
\subsubsection{Audio network}
The audio cues derived from acoustic measurements are projected onto the image plane via the camera model. Therefore, a network similar to the visual network can be used to embed the audio cues into a consistent localization space containing the position context. Further, the audio network contains an additional $\mathit{Conv}$ layer to make the feature dimension consistent with the visual feature.

We design a pre-trained network called GCFNet to learn position representations implicitly. As shown in Figure~\ref{fig-GCFnet}, following the $\mathit{Conv}$ layers, a Multilayer Perceptron (MLP)-based predictor is used to integrate the features into an activation map, which can be formulated as follow:
\begin{equation}
f^z=MLP(Conv^a( \mathbf{R}_{\Omega} )),
\end{equation} 
where $ \mathbf{R}_{\Omega}\in \mathbb{R} ^{m_2 \times H\times  W}$ is short for GCF maps and $f^z\in \mathbb{R} ^{H'\times W'}$ is the output activation map. $Conv^a$ refers to the six-layer audio CNN, whose output vector is the audio feature embedding $f^a\in \mathbb{R} ^{C \times H'\times  W'}$.
$MLP$ refers to the MLP-based predictor with three hidden layers and an average pooling layer. According to the ground-truth (GT) box of the target face $(u,v,h_f,w_f)$, the Gaussian distribution is used to generate a binary map as the pseudo label for GCFNet. 
The pseudo label is calculated by the following formula:
\begin{equation}
L^{GCF}(x,y)=\varepsilon^\lambda (\mathcal{N}(x,y|\mu,\Sigma  )) ,
\end{equation} 
where $(x, y)$ represents the coordinates of the pixel, $\mathcal{N}(\cdot)$ is the two-dimensional Gaussian distribution with mean $\mu =(u,v)^\mathsf{T}$ and covariance $\Sigma  =diag(h_f/\iota,w_f/\iota)$. $\iota$ is a scale factor to adjust the span of the Gaussian surface. $\varepsilon^\lambda(\cdot)$ is a step function that binarizes the label by a fixed threshold $\lambda$. The MSE loss is adopted to train GCFNet.

\begin{figure}[t] 
\centering
\includegraphics[width=1\columnwidth]{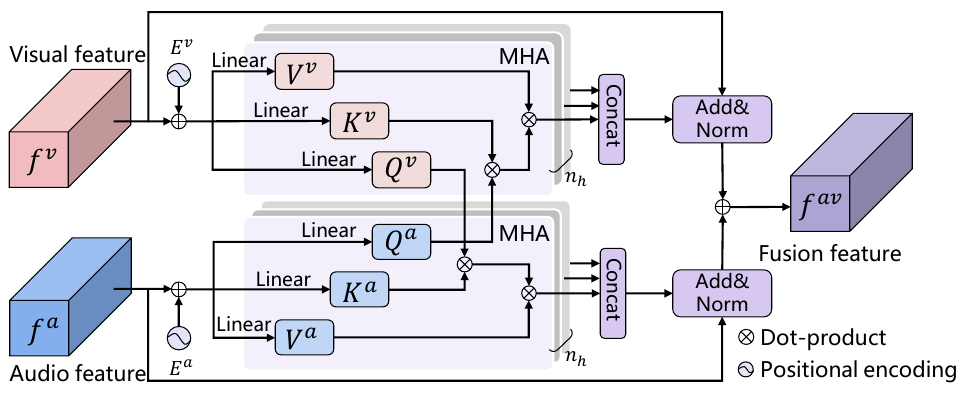} 
\caption{
The structure of cross-modal attention module. The audio and visual features are interact and fused in a multi-head cross-attention mechanism.
}
\label{fig-MHA}
\vspace{-0.1cm}
\end{figure}

\subsection{Cross-Modal Attention}
Inspired by the strong capacity of the transformer model in exploring the interaction among contextual information, features from the visual network and the audio network are fused in a Cross-Modal Attention (CMA) module as shown in Figure~\ref{fig-MHA}. 
The core component of the transformer model is the Multi-Head Attention (MHA) \cite{vaswani2017attention}, which performs the attention function in parallel to consider various attention distributions from multiple representation subspaces.
In each head, the input vector is linearly projected to generate a set of query ($Q_i$), key ($K_i$), and value ($V_i$) with dimension $d_k$, $d_k$ and $d_v$. Attention vectors from multiple heads are concatenated and mapped into a single vector. The mechanism of MHA is summarized as:
\begin{align}
&\mathrm{MHA}(Q,K,V) =Concat(head_1,...,head_{n_h})W^O,\\
&head_i=Softmax(Q_iK_i^{\mathsf{T}}/\sqrt{d_k})V_i,
\end{align}
where $W^O\in \mathbb{R} ^{n_hd_v \times C}$ is the parameter matric. $n_h$ is the number of heads and $d_k=d_v=C/n_h$ in our implementation.


The definition of CMA is similar, where $K$ and $V$ are from the same modality and $Q$ is from another modality, allowing each modality to update its features with external information from the other modality.
CMA integrates multi-modal information by using MHA in the form of residuals. 
The mechanism of CMA is formulated as follow:
\begin{equation}
\begin{split}
f^{av}=& LN(f^{v}+\mathrm{MHA}({Q^a},{K^v},{V^v}) )\\
\oplus & LN(f^{a}+\mathrm{MHA}({Q^v},{K^a},{V^a}) ),
\end{split}
\end{equation} 
where $Q^a$ is the linear transformation of $f^a+E^a$ (the same goes for $Q^v$). $E^a$ is the positional encoding following \cite{chen2021transformer} to supplement spatial positional information. $f^{av}$ is the audio-visual fusion feature.
$\textit{LN}$ denotes the layer normalization.
CMA enables one modality to receive information from another modality, capturing the dependencies of multiple modalities in the global spatial space.
Finally, in the predictor module, the audio-visual fusion feature is fed into an MLP-based regression head to estimate the face coordinates of the target in the current frame.

\subsection{Quality-Aware Module}
In the presence of multiple speakers in a scene, target occlusions and overlapping speech often lead to tracking drift and ID switching. 
To this end, we design a template and search region update/reset strategy for multi-speaker tracking, which is based on a Quality-Aware Module (QAM). 
We construct a light MLP to regress the quality score. 
Based on the difference between the ground truth position and the output of the tracking network, we define the degree of proximity between the two as a training objective for the quality assessment network.
Let $\hat{p}_t$ be the output of the prediction head, the label of the QAM sub-network is defined by a sigmoid function:
\begin{equation}
L^{QAM}=\xi  (e^{\kappa\left \| \hat{p}_t - p^{gt}_t \right \|_2  }+1)^{-1},
\end{equation} 
where $\xi $ and $\kappa$ are hyperparameters to control the intercept and slope. $\left \| \cdot  \right \| _2$ is the Euclidean norm. 
In the inference phase, the regression results of QAM can indicate the quality of the tracking task on the current frame.
The input of QAM is the sum of audio features, visual features, and audio-visual fusion features. We apply two pooling layers and an MLP with a sigmoid activation function in the last layer. The structure of QAM is formulated as follow:
\begin{align}
&f^q = LN(f^a\oplus f^v \oplus f^{av}),\\
&f^{q'}= AvgPool(f^q)+MaxPool(f^q),\\
&q^s = Sigmoid(MLP(f^{q'})),
\end{align}
where $q^s$ is the quality score. We employ the standard MSE loss for the quality-aware module and the prediction head.
\begin{algorithm}
  \caption{Proposed STNet-based tracker}
  \label{alg:tracker}
  \SetAlgoLined
  \KwIn{$t$, $\mathbf{I}_{1:t}$, $\mathbf{S}_{1:t}$, $I_0^{tpl}$, $T$, $m_1$,$m_2$}
  \KwOut{$\hat{p}_{t}$: the predicted target position}
	\While{$t \le T$}{
	    // crop the search region\;
	
		\eIf{$t=1$}{
		$I_t^{tpl}=I_0^{tpl}$\;
		
		$I_t^s=crop(I_t,c(I_0^{tpl}),120\times120)$, $c(\cdot)$ is center point \;
		
		}{
		$I_t^s=crop(I_t,c_{t},120\times120)$\;
		}
		
		// audio-visual processing and localizing \;	
		
		$O_t \gets I_t^s, I_t^{tpl}$ with Eq.\ref{eq:ot}\;
		
		$\mathbf{R}_{\Omega}^{vg}(t,I_t^s), D_{kmax} \gets O_t, I_t^s, s_{t-m_1:t}$ with\textbf{ Algorithm \ref{alg:vg-GCF}}\;
		
		$\hat{p}_{t},q^s \gets STNet(I_t^s, I_t^{tpl}, \mathbf{R}_{\Omega}^{vg}(t,I_t^s))$\;
		
		$\hat{p}_{t}^{3d} \gets \hat{p}_{t}, D_{kmax}$ with Eq. \ref{eq:p3d}\;
		
	 // template and search region update/reset\;
	 
	    $c_{t+1} = \hat{p}_{t}$\;
	    
     \uIf{$q^s \ge \theta_1$}{
          $I_{t+1}^{tpl}= crop(I_t,\hat{p}_{t}, b)$\;
        }
     \uElseIf{$q^s \le  \theta_2$}{
          $I_{t+1}^{tpl}= I_0^{tpl}$\;
          
            \If{$O_t \neq \varnothing$}{$c_{t+1} = o_t$
            }
        }
    \Else{
          $I_{t+1}^{tpl}= I_t^{tpl}$\;
        }
    $t = t + 1$
	}
\end{algorithm}
\vspace{-0.1cm}

\subsection{Tracking with STNet}
To demonstrate the effectiveness and generalization capability of the proposed STNet on the audio-visual speaker tracking task, a simple algorithm is utilized to perform the tracking.
Given a tracking target, an STNet is applied to track a specified single speaker. In the multi-speaker case, multiple STNets perform the tracking task in parallel.
In detail, in each frame, the search area is cropped to a fixed size that is consistent with the size of the input image of the network. The crop center is set to the predicted target position of the previous frame (the center of the original template for the first frame).

In addition, the quality-aware module is added for template and search region update/reset in multi-speaker tracking. 
QAM measures the quality of the outputs of convolutional neural networks and cross-modal attention modules.
Higher quality scores signify more reliable predictions, so updating the template using the current prediction result ensures tracking continuity even if the target appearance changes.
A lower score indicates that the visual observation may be corrupted by target occlusion, or that the audio cues are disturbed by mixed multi-sound source signals, or result in confusing fusion features. 
In this case, we reset the template to the original template that provides highly reliable information and set the crop center of the search region for the next frame according to the visual measurements.
The template update and reset process is summarized as follow:
\begin{equation}
I_{t+1}^{tpl}=
\left\{
\begin{array}{lr}
crop(I_t,\hat{p}_{t}, b)\ \ \  \mathrm{if} \  q^s \ge \theta_1  \\
I_0^{tpl} \ \ \ \ \ \ \ \ \ \ \ \ \ \ \mathrm{if} \   q^s  \le  \theta_2\\
I_{t}^{tpl}\ \ \ \ \ \ \ \ \ \ \ \ \ \ \mathrm{else} ,
\end{array}
\right.
\end{equation}
where $\theta_1$ and $\theta_2$ are threshold parameters used for template update and reset, determined by empirical evaluation.
The superscript $tpl$ denotes the template image patch, and $I_0^{tpl}$ is a pre-defined template in the initial phase. 
The operation $crop(\cdot )$ means to crop the image $I_t$ using $\hat{p}_{t}$ as the center and $b$ as the range. 
When $O_t \neq \varnothing$, $b$ is taken as $b_{lmax}$, otherwise $b$ is set to the scale of $I_0^{tpl}$.
When the window exceeds the image range, the missing part is filled with the average RGB value.

The proposed tracker further executes 3D tracking through the image-3D projection operator provided by a camera model.
After obtaining $\hat{p}_{t}$ in the image plane estimated by the network, with the depth derived from the acoustic measurement, the 3D coordinate is calculated as:
\begin{equation}
\label{eq:p3d}
\hat{p}_{t}^{3d}=\Phi (\hat{p}_{t};D_{kmax}),
\end{equation}
where $D_{kmax}$ is the depth where the peak of vgGCF maps is located and ${kmax}$ is the corresponding index.
$\Phi$ is the image-3D projection operator, which requires camera calibration parameters. The complete tracking process of the STNet-based Tracker (STNT) is summarized in Algorithm \ref{alg:tracker}.

\section{Experiments and Discussions}
In this section, the proposed tracker is compared with state-of-the-art methods and uni-modal methods on the AV16.3 \cite{AV163} dataset. First, the dataset, evaluation metrics, and experimental setups are described in detail. Then, the comparative results and the effectiveness of each component are discussed. Finally, the visualization results are presented to further demonstrate the effectiveness of the proposed method.

\subsection{Dataset}
\textbf{AV16.3} \cite{AV163} is an audio-visual corpus containing indoor multispeaker recordings designed to test algorithms for audio-only, video-only, and audio-visual speaker localization and tracking. 
In the provided sequences, 1-3 participants speak in the conference room while moving and displaying various actions, which are simultaneously recorded by three calibrated cameras and two circular eight-element microphone arrays.
Audio data is captured at a sample rate of $16kHz$ by microphone arrays mounted on the table. 
The images are collected at $25Hz$ by monocular color cameras installed in three corners of the room, with resolution as $288 \times 360$ pixels. 

In each experiment, data captured by two microphone arrays and one of the three cameras are used sequentially to validate the effectiveness of the algorithm at different viewpoints.
For training, we use the single-speaker sequences $seq01,02,03$. For evaluation, we test the same sequences specified in \cite{VK2015TMM, VK2016TMM, Qian2018ICASSP, qian2019TMM, Qian2022TMM,lyd2022AAAI}, using $seq08, 11, 12$ for Single Object
Tracking (SOT) and $seq24, 25, 30$ for Multiple Object Tracking (MOT).

\textbf{CAV3D} \cite{qian2019TMM} is an audio-visual corpus of speaker tracking collected by a co-located sensor platform. The dataset contains multi-speaker data collected in a room of size ($4.77 \times 5.95 \times 4.5, m$). The co-located sensor platform placed on the conference table consists of a camera and an 8-element circular microphone array. The 8-channel audio signal is recorded at 96 kHz and the $768 \times 1024$ pixel video signal is recorded at 15 frames per second with a field of view of 90°. In addition, four corner cameras are configured for 3D trajectory calibration. The dataset consists of 20 sequences divided into three types of subsets, CAV3D-SOT (9 single-speaker sequences), CAV3D-SOT2 (6 sequences with one speaker and one non-vocalized participant), and CAV3D-MOT (5 multi-speaker sequences). Compared to AV16.3, the scenes in the CAV3D dataset are more complex, including scenes with occlusion, non-frontal views, entering/exiting the camera field of view, and long periods of silence. We collect data on 9 sequences in the CAV3D-SOT subset for model training.
\begin{figure}[t] 
\centering
\includegraphics[width=1\columnwidth]{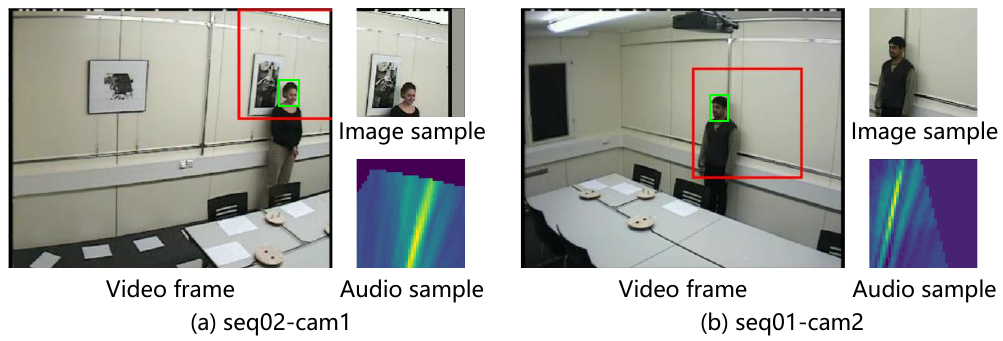} 
\caption{
Illustration of audio-visual sample pairs in training. The image samples are randomly sampled around GT (green) and are padded to $120 \times 120$.
The audio sample is a set of vgGCF maps (only one of which is shown).
}
\label{fig-sample}
\end{figure}
\subsection{Metrics}
\textbf{Mean Absolute Error (MAE)} is a tracking performance metric that is estimated by calculating the Euclidean distance between the predicted positions and ground-truth positions. 
This metric is widely used because it explicitly measures the accuracy of target localization for performance comparison. MAE is defined as follow:
\begin{equation}
\mathrm{MAE}=\frac{1}{TN} \sum_{t=1}^{T} \sum_{n=1}^{N}\left \| \hat{p}_{t,n} - p^{gt}_{t,n} \right \| _2,
\end{equation} 
where $T$ is the total number of frames and $N$ is the total number of objects in each frame. MAE can be evaluated in the image plane (in $pixels$) or 3D world coordinate (in $m$).

\textbf{MOT Metric} includes MOT Precision (MOTP) and MOT Accuracy (MOTA) \cite{mot-metric}. Unlike MAE, MOTP only calculates the errors in successfully matched frames, avoiding the considerable impact of huge errors caused by target losses.
MOTA provides an intuitive measure of the tracker's performance in detecting objects and maintaining trajectories, which is independent of the accuracy of object position estimation.
MOTA considers the number of false decisions made by the tracker, including FP (errors greater than a threshold), FN (missing targets), and IDS (speaker identity being switched):
\begin{equation}
\mathrm{MOTA} = 100 \times (1-\frac{ {\textstyle \sum_{t}(FP_t+FN_t+IDS_t)} }{\sum_{t}n_t^{gt}} ),
\end{equation} 
where $n_t^{gt}$ denotes the ground-truth number at time $t$. With the same settings as in \cite{Qian2022TMM}, we record the results with errors exceeding 1/15 of the image diagonal size as FP.
\begin{table*}[]
\footnotesize
\centering
\caption{MOT experimental results for uni-modal methods and state-of-the-art audio-visual methods on the AV16.3 dataset. All metrics are calculated on the image plane. * indicates that the method uses DOA with annotations as a priori.
	}  
\label{table:MOT}
 \setlength{\tabcolsep}{3.05pt}
 \renewcommand{\arraystretch}{1.2}
\begin{tabular}{cc|ccccccc|ccccccc|cccccc}
\toprule 
\multicolumn{2}{c|}{Sequences} & \multicolumn{7}{c|}{MAE $\downarrow$}                                              & \multicolumn{7}{c|}{MOTA$ \uparrow$}                                                   & \multicolumn{6}{c}{MOTP $\downarrow$}                                \\ \hline
\multicolumn{1}{c|}{seq} & cam & AO    & VO    & \cite{VK2015TMM}* & \cite{VK2016TMM} & \cite{qian2019TMM}& \cite{ImprovedGCF} & Ours & AO    & VO    & \cite{VK2015TMM}* & \cite{VK2016TMM} & \cite{qian2019TMM} & \cite{Qian2022TMM} & Ours   & AO    & VO    & \cite{VK2015TMM}* & \cite{VK2016TMM} & \cite{qian2019TMM} & Ours          \\ \hline
\multicolumn{1}{c|}{}    & 1   & 51.36 & 30.24 & 11.27   & 18.43  & 4.45          & 12.10         & \textbf{4.26}  & 48.96 & 88.17 & 79.61     & 78.57     & 97.25       & 89.40          & \textbf{97.85} & 11.64 & 7.56  & 11.27     & 10.48     & 4.42        & \textbf{4.16} \\
\multicolumn{1}{c|}{24}  & 2   & 42.25 & 28.64 & 10.24   & 12.22  & 38.21         & 11.90         & \textbf{5.02}  & 61.65 & 64.15 & 85.48     & 79.58     & 61.02       & 91.00          & \textbf{94.16} & 9.28  & 6.84  & 10.24     & 9.11      & 6.09        & \textbf{4.24} \\
\multicolumn{1}{c|}{}    & 3   & 48.22 & 30.52 & 10.43   & 14.75  & 34.55         & \textbf{8.30} & 25.49          & 46.34 & 65.04 & 77.79     & 59.46     & 57.71       & \textbf{91.00} & 61.36          & 10.26 & 9.66  & 10.43     & 10.78     & 8.69        & \textbf{4.93} \\ \hline
\multicolumn{1}{c|}{}    & 1   & 78.74 & 25.23 & 15.41   & 15.55  & 12.33         & \textbf{9.80} & 20.79          & 41.83 & 58.02 & 69.25     & 82.18     & 70.57       & \textbf{94.80} & 64.91          & 19.08 & 10.87 & 15.33     & 14.32     & 8.96        & \textbf{6.25} \\
\multicolumn{1}{c|}{25}  & 2   & 37.27 & 31.12 & 11.14   & 19.12  & 9.12          & 10.90         & \textbf{6.51}  & 63.68 & 75.70 & 79.08     & 64.44     & 84.37       & 85.90          & \textbf{90.49} & 8.12  & 8.82  & 9.06      & 8.14      & 7.94        & \textbf{5.13} \\
\multicolumn{1}{c|}{}    & 3   & 53.93 & 41.43 & 9.70    & 12.40  & \textbf{9.15} & 12.00         & 10.16          & 48.98 & 56.28 & 82.25     & 80.62     & 87.55       & \textbf{91.20} & 82.72          & 15.32 & 9.85  & 9.70      & 11.30     & 6.85        & \textbf{6.11} \\ \hline
\multicolumn{1}{c|}{}    & 1   & 45.83 & 52.62 & 24.72   & 16.91  & \textbf{7.23} & 11.30         & 16.33          & 53.69 & 45.74 & 62.00     & 69.85     & \textbf{88.35}       & 40.80          & 70.45          & 12.27 & 11.22 & 11.69     & 12.48     & 6.92        & \textbf{6.18} \\
\multicolumn{1}{c|}{30}  & 2   & 86.12 & 32.11 & 9.12    & 11.42  & 6.69          & 7.80          & \textbf{4.85}  & 35.60 & 48.58 & 85.15     & 86.67     & 97.15       & 82.70          & \textbf{99.53} & 19.04 & 8.99  & 9.12      & 10.49     & 6.62        & \textbf{4.63} \\
\multicolumn{1}{c|}{}    & 3   & 79.38 & 42.37 & 10.31   & 11.30  & 5.16          & 19.50         & \textbf{4.72}  & 39.12 & 58.65 & 74.65     & 58.41     & 96.50       & 64.70          & \textbf{99.42} & 15.16 & 6.90  & 10.31     & 13.01     & 5.12        & \textbf{4.72} \\ \hline
\multicolumn{2}{c|}{Average}   & 58.12 & 34.92 & 12.48   & 14.68  & 14.10         & 11.51         & \textbf{10.90} & 48.87 & 62.26 & 76.69     & 73.31     & 82.27       & 81.28          & \textbf{84.10} & 13.35 & 8.97  & 10.79     & 11.12     & 6.85        & \textbf{5.15}\\ 
\bottomrule 
\end{tabular}
\end{table*}

\begin{table*}[h]
\centering
\footnotesize
\caption{SOT Experimental results for uni-modal methods and state-of-the-art audio-visual methods on the AV16.3 dataset. 2D MAE is evaluated on the image plane in $pixels$, and 3D MAE is evaluated in the 3D world coordinate in $m$.
	}  
\label{table:SOT}
  \setlength{\tabcolsep}{6.75pt}
  \renewcommand{\arraystretch}{1.2}
\begin{tabular}{cc|ccccccc|cccc}
\toprule 
\multicolumn{2}{c|}{Sequences} & \multicolumn{7}{c|}{2D MAE ($pixels$) $\downarrow$}         & \multicolumn{4}{c}{3D MAE ($m$) $\downarrow$} \\ \hline
\multicolumn{1}{c|}{seq} & cam & AO    & \multicolumn{1}{c|}{VO}    & \cite{VK2015TMM}  & \cite{Qian2018ICASSP} & \cite{qian2019TMM} & \cite{lyd2022AAAI} & Ours & AO     & \cite{qian2019TMM}   & \cite{Qian2018ICASSP}   & Ours  \\ \hline
\multicolumn{1}{c|}{}    & 1   & 25.82 & \multicolumn{1}{c|}{15.85} & 10.68 & 4.31   & 4.22   & 3.51 & \textbf{3.33} & 0.18   & 0.16  & 0.12  & \textbf{0.12}  \\
\multicolumn{1}{c|}{08}  & 2   & 14.09 & \multicolumn{1}{c|}{10.55} & 8.12  & 4.66   & 6.32   & 3.69 & \textbf{3.23} & 0.21   & 0.14  & \textbf{0.11}  & 0.13  \\
\multicolumn{1}{c|}{}    & 3   & 21.09 & \multicolumn{1}{c|}{9.48}  & 9.63  & 5.34   & 6.32   & 4.13 & \textbf{3.86} & 0.19   & 0.10  & \textbf{0.09}  & 0.13  \\ \hline
\multicolumn{1}{c|}{}    & 1   & 22.28 & \multicolumn{1}{c|}{11.81} & 16.51 & 8.15   & 9.27   & 6.12 & \textbf{4.30} & 0.25   & 0.23  & 0.33  & \textbf{0.16}  \\
\multicolumn{1}{c|}{11}  & 2   & 19.42 & \multicolumn{1}{c|}{10.98} & 15.16 & 7.48   & 9.79   & 4.25 & \textbf{3.30} & 0.24   & 0.12  & 0.14  & \textbf{0.10}  \\
\multicolumn{1}{c|}{}    & 3   & 21.81 & \multicolumn{1}{c|}{12.55} & 15.45 & 6.64   & 7.72   & 5.84 & \textbf{3.39} & 0.26   & 0.14  & 0.12  & \textbf{0.11}  \\ \hline
\multicolumn{1}{c|}{}    & 1   & 35.99 & \multicolumn{1}{c|}{17.06} & 12.63 & 6.86   & 10.24  & 5.65 & \textbf{3.74} & 0.57   & 0.21  & 0.26  & \textbf{0.19}  \\
\multicolumn{1}{c|}{12}  & 2   & 21.48 & \multicolumn{1}{c|}{38.50} & 10.48 & 10.67  & 19.21  & 4.89 & \textbf{3.61} & 0.46   & 0.29  & 0.17  & \textbf{0.17} \\
\multicolumn{1}{c|}{}    & 3   & 30.14 & \multicolumn{1}{c|}{57.36} & 15.21 & 9.71   & 16.01  & 3.38 & \textbf{3.00} & 0.53   & 0.32  & \textbf{0.20}  & 0.21  \\ \hline
\multicolumn{2}{c|}{Average}   & 23.57 & \multicolumn{1}{c|}{20.46} & 12.65 & 7.09   & 9.90   & 4.61 & \textbf{3.53} & 0.32   & 0.19  & 0.17  & \textbf{0.14}  \\
\bottomrule 
\end{tabular}
\end{table*}
\subsection{Implementation Details}
In the audio branch, the speech signal from circular microphone arrays is enframed to $40ms$ by a Hamming window with 1/2 frame shift.
The parameters for calculating vgGCF are set as $d=9$, $m_1=15$, $m_2=3$.
The 2D sample points are separated by 3 pixels. 
According to the actual situation of the room configuration of the two datasets, the 3D sampling points located outside the room range and below the tabletop are removed.
A pinhole camera model is built using camera calibration parameters provided by the datasets to implement the image-3D projection.
In the visual branch, the search images are resized to 3 sizes, $2.5\times$, $3.3\times$, and $4.2\times$, to handle scale variations.
The confidence threshold $\theta^v$ for visual observation is set to 0.3 on AV16.3 and 0.35 on CAV3D.
In the training phase, we collect 44,958 sets of audio-visual sample pairs from the training sequences in AV16.3, and 73,256 pairs on CAV3D-SOT.
We capture $120\times 120$ image samples from AV16.3 and $320\times 320$ image samples from CAV3D in a data enhancement manner of surrounding cropping, as shown in Figure~\ref{fig-sample}.
The image patches
are randomly cropped around the ground-truth bounding box and completely contain the ground-truth box.
The audio samples are vgGCF maps extracted in the corresponding sampling area.
The parameters to build $L^{GCF}$ are set to $ \iota=2.5$, $\lambda=0.5$. 
The parameters to construct $L^{QAM}$ are set to $ \xi=2$, $\kappa=0.05$.
In the CAM module, $n_h$ is set to 8.
In the tracking test, the thresholds that control update and reset are set to $\theta_1=0.98$, $\theta_2=0.94$ on AV16.3 and $\theta_1=0.985$, $\theta_2=0.945$ on CAV3D.

For model training, the STNet is trained using pre-training with fine-tuning. 
In the visual branch, the Siamese-based feature extraction network is pre-trained on 214 sequences on the GOT-10k \cite{Got-10k} dataset. We use the Adam optimizer with a learning rate of $10^{-3}$ and a batch size of 32. 
In the audio branch, the GCFNet model is pre-trained using data from random surrounding cropping on the training sequences in AV16.3 and CAV3D. 
We use the Adam optimizer with the learning rate of $10^{-4}$ and a batch size of 16. The early stopping strategy is employed to prevent overfitting. 
Subsequently, the fusion module and QAM block are added after the uni-modal branch to train the complete network with an optimizer of SGD, a learning rate of $10^{-4}$, and a batch size of 16.
The experiments are conducted on the PyTorch framework with one NVIDIA RTX 2080 Ti GPU.  The code is publicly available at \href{https://github.com/liyidi/STNet}{https://github.com/liyidi/STNet}.

\subsection{Comparisons with State-of-the-art Results}
Our method is compared with uni-modal methods and previous state-of-the-art Audio-Visual (AV) methods. 
The results of MOT and SOT on two datasets are shown in Table \ref{table:MOT}, \ref{table:SOT} and \ref{table: CAV3D}, respectively.
Where the Audio-Only (AO) and Visual-Only (VO) methods are implemented based on stGCF and the visual network described in Sec.~\ref{AO} and Sec. ~\ref{VO}.
We reproduce the methods of \cite{VK2015TMM,VK2016TMM,qian2019TMM,lyd2022AAAI} by running their source codes. 
The results of \cite{Qian2018ICASSP}, \cite{Qian2022TMM} and \cite{ImprovedGCF} are reported in their papers.
Note that \cite{VK2015TMM}* uses the annotated DOA as a priori to avoid huge errors due to incorrect correspondence of person ID.

The excellent performance of AV trackers compared to uni-modal approaches indicates the effectiveness of audio-visual fusion.
On the AV16.3 dataset, our method outperforms the state-of-the-art methods on SOT and MOT under all metrics.
Table \ref{table:SOT} shows that the proposed STNT outperforms the filter-based methods \cite{VK2015TMM, Qian2018ICASSP, qian2019TMM} considerably in tracking accuracy and achieves a lower MAE (3.53 pixels) than MPT presented in \cite{lyd2022AAAI}.
Where MPT also utilizes a deep neural network for multi-modal feature fusion, demonstrating the superiority of deep learning methods for object tracking and modality perception.
MPT employs an additional particle filter algorithm with higher computational costs to further improve the tracking accuracy. Whereas we only use a simplistic tracking strategy to obtain more accurate results. 
Table \ref{table:MOT} shows that the average results of our STNT in MOT experiments outperform the other methods. 
Large MAEs in MOT experiments mainly come from ID assignment errors due to the occlusion between walking speakers.
\cite{VK2016TMM} uses PHD filtering and \cite{qian2019TMM,Qian2022TMM} employ additional data association algorithms for observation-to-track assignments.
\cite{ImprovedGCF} is competitive in multi-speaker tracking because of the usage of a separation-before-localization method, i.e., separating the multi-speaker audio mixture into multiple single-speaker audios, and then calculating the GCF features of the individual sources for position estimation. However, its tracking effectiveness is limited by the accuracy of the speech separation technique.

In the proposed tracking framework, benefiting from the update/reset strategy based on QAM, STNT is able to re-detect the targets after they are lost, ensuring the continuity of tracking while simplifying the complicated association problem between multi-modal observations and multiple speakers.
Without considering tracking configuration errors, the MOTP calculated on successfully tracked frames measures the precision of the tracking system.
As shown in Table \ref{table:MOT}, the average MOTP (5.15 $pixels$) for STNT shows a competitive performance, displaying the advantages of STNet in the speaker tracking task.
Moreover, we derive the 3D trajectory in the world coordinate in the SOT test. 
As shown in Table \ref{table:SOT}, the average 3D error of STNT is 0.14 $m$, which outperforms the AO method and comparison methods.
The precise 3D tracking relies on the accurate localization of STNet and the depth estimation of the acoustic measurement. 

\begin{table}[t]
\footnotesize
\centering
\caption{Experimental results for uni-modal methods and state-of-the-art audio-visual methods on the CAV3D dataset. All metrics are calculated on the image plane.
	}  
\label{table: CAV3D}
\setlength{\tabcolsep}{7pt}
\renewcommand{\arraystretch}{1.2}
\begin{tabular}{c|c|ccccc}
\toprule 
Sequences  & 2D   & AO    & VO   & \cite{lyd2022AAAI}  & \cite{qian2019TMM}           & Ours*         \\ \hline
\multirow{2}{*}{SOT2} & MAE $\downarrow$  & 22.47 & 9.88 & 9.38  & 7.70           & \textbf{7.26} \\
     & MAE* $\downarrow$ & 9.28  & 5.69 & 5.65  & 5.30           & \textbf{5.16} \\ \hline
\multirow{2}{*}{MOT}  & MAE $\downarrow$  & 46.54 & 20.2 & - & \textbf{10.10} & 12.38         \\
     & MAE* $\downarrow$ & 9.76  & 5.84 & -  & 4.90           & \textbf{4.82} \\ \bottomrule 
\end{tabular}
\end{table}

The experimental results for the image plane on the CAV3D dataset are shown in Table \ref{table: CAV3D}. MAE* in \cite{qian2019TMM} is introduced to denote the MAE computed only on frames where tracking is successful.
In the CAVSD-MOT test sequences, the speaker may go out of the camera's field of view.
Our* indicates that our method uses an additional front-end target detector to capture the speaker image that re-enters the frame as the initial template for the tracker, which is a limitation of our method. 
Note that MOT results for \cite{lyd2022AAAI} are not listed because it is not applicable to multi-target tasks.
\cite{qian2019TMM} uses a data association algorithm based on a greedy strategy and therefore performs well on MOT sequences.
The superior performance of STNT on SOT2 demonstrates the tracking algorithm's generalizability across different datasets.

\subsection{Ablation Study}
\begin{figure*}[t] 
\centering
\includegraphics[width=2\columnwidth]{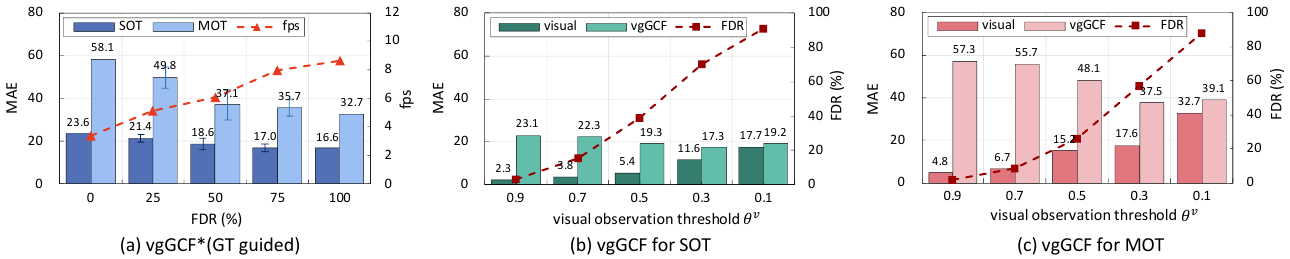} 
\caption{
Illustration of the guiding effect of visual information. (a) vgGCF* is guided by the randomly selected GT face bounding boxes. fps: frame per second; (b) vgGCF for SOT and (c) vgGCF for MOT are guided by visual observations derived from the visual network. The folded lines indicate the Face Detection Rate (FDR) at different visual thresholds. All errors are calculated in the image plane.
}
\label{fig-exp-GCFFDR}
\vspace{-0.1cm}
\end{figure*}

\textbf{Acoustic Measurement.}
The proposed vgGCF is compared with the general GCF (peaks in 3D space are projected onto the image plane) and stGCF in experiments of SSL and tracking.
The significant improvements in Table~\ref{table:SSL} and Table ~\ref{table:GCFnet} demonstrate the effectiveness of the proposed acoustic measurement framework. 
With the addition of visual guidance, the accuracy of sound source localization is further improved, primarily in the case of multiple sources.
The stGCF method focuses only on the peaks in the sampling space and thus cannot distinguish multiple active candidate sources. 
In contrast, the search space in vgGCF is narrowed down to the face region, thus avoiding the interference of multiple sound sources.
In Figure~\ref{fig-exp-GCFFDR}, we evaluate the SSL results using visual information with different Face Detection Rate (FDR).
We test 0\%-100\% of randomly selected GT faces to achieve the ideal vgGCF*.
The vertical line at the top of the bar in Figure~\ref{fig-exp-GCFFDR} (a) indicates the standard deviation over ten repetitions.
As the FDR increases, the calculation amount of each frame decreases (fps increases). 
This is due to the reduction of the distribution of spatial sampling points from the global area to the face detection box.
In the practical test, the visual observations are derived from the visual network, and the results are shown in Figure~\ref{fig-exp-GCFFDR} (b-c).
As the threshold value decreases, the number of visual observations larger than the threshold value increase, and the corresponding detection error increases.
Incorrect visual information leads to an increased SSL error. In the tracking implementation, the visual observation threshold is set to the optimal value of 0.3.
\begin{table}[t]
\footnotesize
\centering
\caption{Experimental results of sound source localization using different acoustic measurements and GCFNet.
	}  
\label{table:SSL}
\setlength{\tabcolsep}{8pt}
\renewcommand{\arraystretch}{1.2}
\begin{tabular}{c|c|ccc}
\toprule 
\multirow{2}{*}{method} & SOT            & \multicolumn{3}{c}{MOT}                                                                   \\ \cline{2-5}
                        & MAE $\downarrow$            & \multicolumn{1}{c|}{MAE $\downarrow$}            & \multicolumn{1}{c|}{MOTA $\uparrow$}           & MOTP $\downarrow$          \\ \hline
GCF                     & 74.28          & \multicolumn{1}{c|}{-}              & \multicolumn{1}{c|}{-}              & -             \\
stGCF                   & 23.57          & \multicolumn{1}{c|}{58.12}          & \multicolumn{1}{c|}{48.87}          & 13.35         \\
vgGCF                   & 17.34          & \multicolumn{1}{c|}{37.53}          & \multicolumn{1}{c|}{60.45}          & 9.58          \\
stGCF+GCFNet            & 18.26          & \multicolumn{1}{c|}{39.12}          & \multicolumn{1}{c|}{56.68}          & 11.43         \\
vgGCF+GCFNet            & \textbf{15.38} & \multicolumn{1}{c|}{\textbf{28.72}} & \multicolumn{1}{c|}{\textbf{68.21}} & \textbf{8.24} \\ 
\bottomrule 
\end{tabular}
\end{table}

\begin{table}[t]
\footnotesize
\centering
\caption{Ablation study on pre-trained GCFNet in model training and acoustic measurement (AM) in tracking tests.
	}  
\label{table:GCFnet}
\setlength{\tabcolsep}{4.5pt}{
\renewcommand{\arraystretch}{1.2}
\begin{tabular}{clc|c|ccc}
\toprule 
\multicolumn{3}{c|}{method}                               & SOT           & \multicolumn{3}{c}{MOT}                                                                  \\ \hline
\multicolumn{2}{c|}{model}                   & AM         & MAE $\downarrow$           & \multicolumn{1}{c|}{MAE $\downarrow$}           & \multicolumn{1}{c|}{MOTA $\uparrow$ }          & MOTP $\downarrow$          \\ \hline
\multicolumn{2}{c|}{STNet without}                & stGCF      & 7.12          & \multicolumn{1}{c|}{34.62}         & \multicolumn{1}{c|}{71.46}          & 8.83          \\
\multicolumn{2}{c|}{pre-trained GCFNet} & vgGCF      & 5.34          & \multicolumn{1}{c|}{14.46}         & \multicolumn{1}{c|}{80.25}          & 6.92          \\ \cline{2-2}
\multicolumn{2}{c|}{STNet with}                & stGCF      & 5.48          & \multicolumn{1}{c|}{28.15}         & \multicolumn{1}{c|}{73.98}          & 7.57          \\
\multicolumn{2}{c|}{pre-trained GCFNet}    & vgGCF      & \textbf{3.53} & \multicolumn{1}{c|}{\textbf{10.90}} & \multicolumn{1}{c|}{\textbf{84.10}} & \textbf{5.15} \\ 
\bottomrule 
\end{tabular}}
\end{table}

\begin{figure}[t] 
\centering
\includegraphics[width=1\columnwidth]{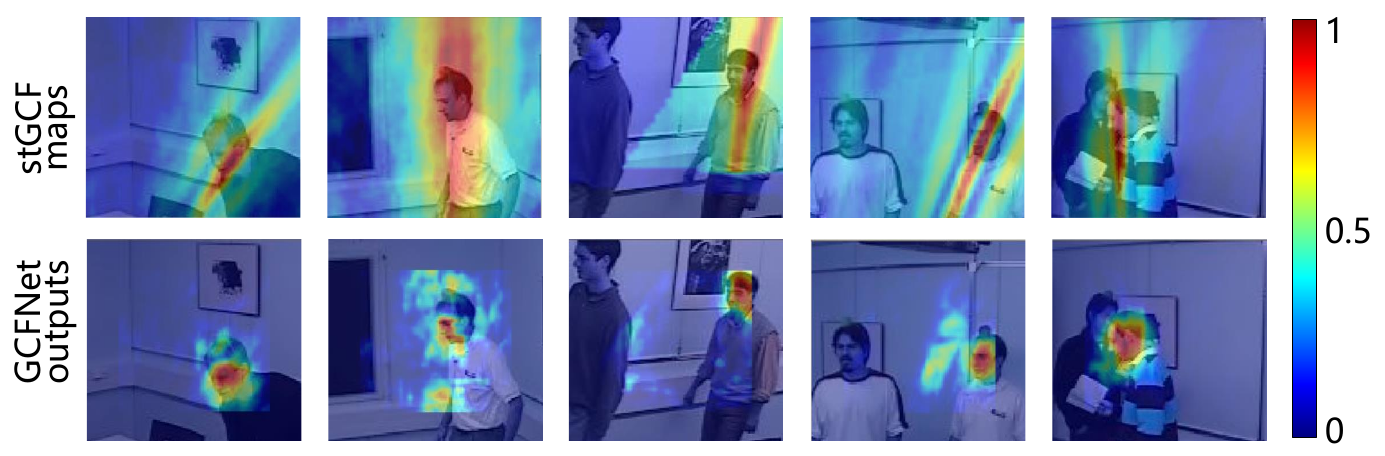} 
\caption{
The visualization results of the GCFNet. First line: input stGCF maps; Second line: GCFNet outputs. The background image is only used to show the speaker position and pose without providing visual information.
}
\label{fig-exp-GCFnet}
\vspace{-0.1cm}
\end{figure}

\textbf{Audio network.}
We quantify the SSL performance of the pre-trained GCFNet and its effectiveness in the proposed STNT.
The last two rows of Table \ref{table:SSL} show that the localization results processed by GCFNet are more accurate than the original stGCF and vgGCF maps.
Figure~\ref{fig-exp-GCFnet} displays the visualization of the network inputs and outputs.
In the acoustic maps, regions near the sound source are shown highlighted, indicating a high probability of sound source presence.
Since the speaking voice is directional, high response values on the line from the speaker's mouth to the microphone array are projected onto the image plane, making the results of source localization ambiguous.
In the output maps of GCFNet, the highlighted areas are visibly narrowed and concentrated on the speaker's face because the supervision at the training phase comes from the pseudo-labels generated by the face bounding box.
The ablation experiments in Table \ref{table:GCFnet} show that using the pre-trained GCFNet to initialize the parameters of audio CNN module in STNet yields improved performance with error reductions of 1.81 and 3.56 $pixels$ on SOT and MOT.
The representations learned by GCFNet with respect to sound source locations are migrated to STNet, demonstrating the effectiveness of the training paradigm.

\begin{figure*}[t] 
\centering
\includegraphics[width=2.1\columnwidth]{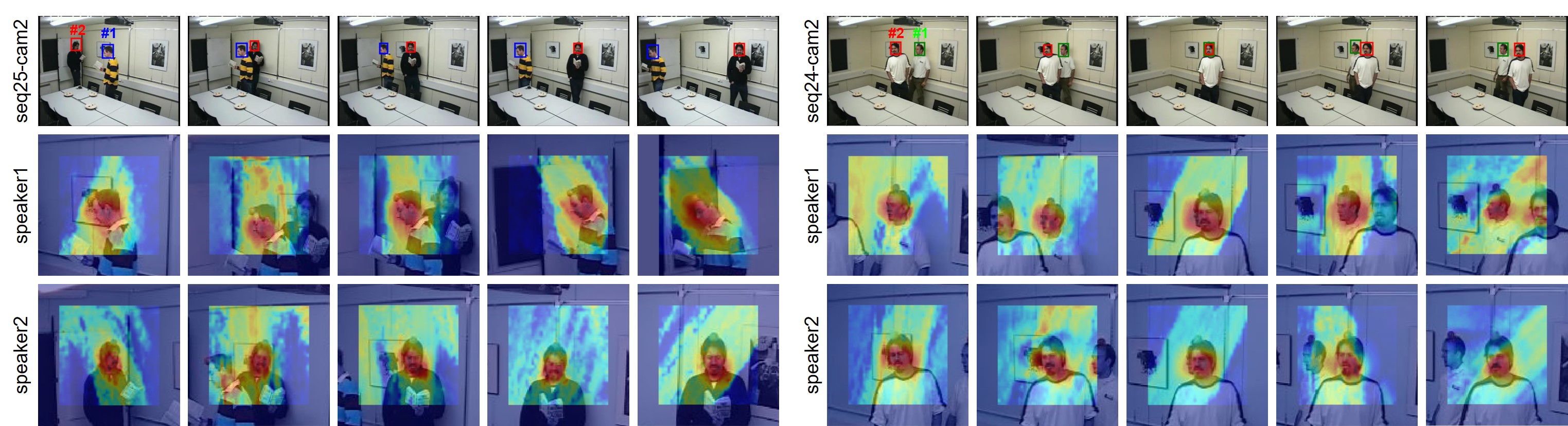} 
\caption{
Visualization of the fusion feature maps output by the cross-modal attention module for two representative sequences. The first row presents the video frames with the tracking result boxes of STNT. The second and third rows show the fusion features in the tracker corresponding to each speaker, respectively.
}
\label{fig-fuisonmap}
\vspace{-0.15cm}
\end{figure*}

\begin{figure}[t] 
\centering
\includegraphics[width=1\columnwidth]{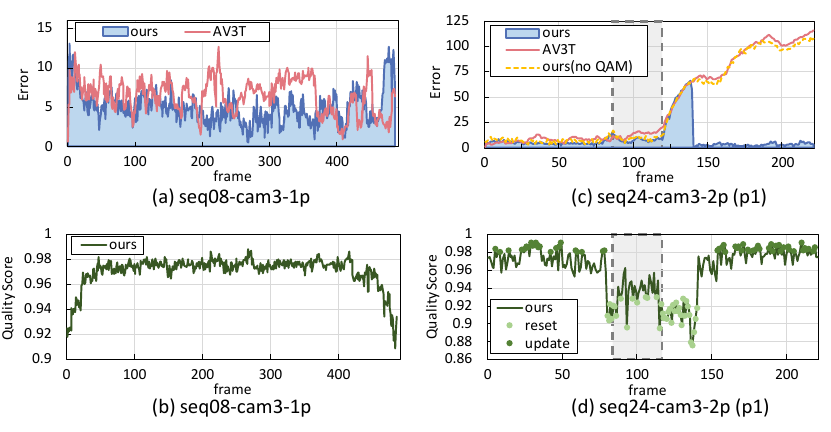}
\caption{
Error and quality score for each frame in seq08-cam3 (a-b) and seq24-cam3 (c-d). AV3T is the tracker proposed in \cite{qian2019TMM}. The gray area indicates that the target is occluded.
}
\label{fig-exp-QAM}
\vspace{-0.35cm}
\end{figure}

\begin{table}[t]
\footnotesize
\centering
\caption{Ablation study on fusion method. The proposed CMA is compared with the alternative solutions.
	}  
\label{table:Fusion Method}
\setlength{\tabcolsep}{7.5pt}
\renewcommand{\arraystretch}{1.2}
\begin{tabular}{c|c|ccc}
\toprule 
\multirow{2}{*}{Fusion Method} & SOT           & \multicolumn{3}{c}{MOT}                                                                  \\ \cline{2-5} 
                               & MAE $\downarrow$           & \multicolumn{1}{c|}{MAE $\downarrow$}           & \multicolumn{1}{c|}{MOTA $\uparrow$}           & MOTP $\downarrow$        \\ \hline
Multiplicative                 & 8.95          & \multicolumn{1}{c|}{33.28}         & \multicolumn{1}{c|}{61.06}          & 11.54         \\
Additive                       & 10.54         & \multicolumn{1}{c|}{31.39}         & \multicolumn{1}{c|}{63.45}          & 12.66         \\
Concatenate                    & 8.68          & \multicolumn{1}{c|}{28.72}         & \multicolumn{1}{c|}{70.12}          & 9.69          \\
Concatenate+CNN                & 6.59          & \multicolumn{1}{c|}{30.42}         & \multicolumn{1}{c|}{68.45}          & 7.63          \\ 
Self-attention+CMA             & 4.82         & \multicolumn{1}{c|}{14.27}         & \multicolumn{1}{c|}{79.43}          & 7.14          \\ 
CMAF (\cite{cmaf})    &4.75 & \multicolumn{1}{c|}{14.18} & \multicolumn{1}{c|}{80.25} & 7.02  \\ 
CMA(Ours)                      & \textbf{3.53} & \multicolumn{1}{c|}{\textbf{10.90}} & \multicolumn{1}{c|}{\textbf{84.10}} & \textbf{5.15} \\ 
\bottomrule 
\end{tabular}
\vspace{-0.2cm}
\end{table}
\textbf{Fusion Strategy.}
To quantify the effectiveness of the cross-modal attention module as a fusion strategy, the proposed CMA is compared with six alternative solutions.
In the additive (multiplicative) fusion strategy, visual and audio features with the same dimension are processed using the element-wise addition (product) operation.
In the concatenate strategy, the visual and audio features are connected along the channel dimension.
In the "Concatenate+CNN" scheme, the connected fusion features are fed into a residual block containing two CNNs.
In the "Self-attention+CMA" scheme, we add the self-attention module for each modality before CMA.
In addition, we replace the CMA module with the Cross-Modal Attentive Fusion (CMAF) mechanism proposed in \cite{cmaf}.
All alternatives show promising performance on SOT, demonstrating the effectiveness of the proposed framework.
Additive fusion has stronger noise immunity, while multiplicative fusion shows sharper peaks in the probability density distribution.
The noise accumulation leads to the appearance of pseudo-peaks, which increase the false positives of multiplicative fusion on MOT sequences.
The concatenation operation preserves more information about the original features and avoids the confusion of semantic information between different channels.
The parameter size of the alternative CNN block is 4.7M. 
In contrast, CMA has fewer parameters (0.28M).
Compared to CMA, CMAF incorporates a self-attention module in parallel for exploiting intra-modal temporal correlations. 
Using CMA alone works better than CMAF and "Self-attention+CMA" which combines self-attention due to the fact that long-range dependencies in a single modality captured by self-attention are not suitable for this end-to-end tracking framework, where the feature embedding input to fusion module, vgGCF, already takes into account time-dependent information.
Nevertheless, the improvement brought by the cross-attention mechanism is pronounced, as it introduces the stream from the audio (visual) modality into the visual (audio) modality via the attention mechanism.
By focusing on global contexts, CMA implicitly models the interactions across multiple modalities to better exploit the advantages of audio-visual fusion.

\textbf{Quality-Aware Module.}
To demonstrate the effect of the proposed QAM and update/reset strategy, we plot the line graphs of the error and quality scores. 
Figure~\ref{fig-exp-QAM} (a-b) shows the results on a single-speaker sequence. 
The update/reset strategy is not applied for SOT, and the quality scores output here are only used to display the results of QAM.
Excluding the occlusion interference and the improvement of the update/reset strategy, the quality score on the single-speaker sequence can more intuitively represent the effect of QAM.
Figure~\ref{fig-exp-QAM} (a) displays that STNT makes larger errors on the beginning and end frames of the sequence, showing two peaks on the line graph.
In contrast, the quality scores are smaller at the beginning and end frames.
The opposite trend proves the effectiveness of QAM.
Ablation experiments of QAM are performed on a multi-target sequence.
The comparison method does not utilize the QAM-based update/reset strategy.
The gray area in Figure~\ref{fig-exp-QAM} (c)(d) indicates that the target is occluded.
The light (dark)-colored dots in Figure~\ref{fig-exp-QAM} (d) denote the reset (updated) frames.
The error lines in Figure~\ref{fig-exp-QAM} (c) illustrate that the trackers in the comparison methods lost the speaker after he passed through occlusion, while our method rediscovers the target after a short period of misjudgment. 

\begin{figure}[t] 
\centering
\includegraphics[width=1\columnwidth]{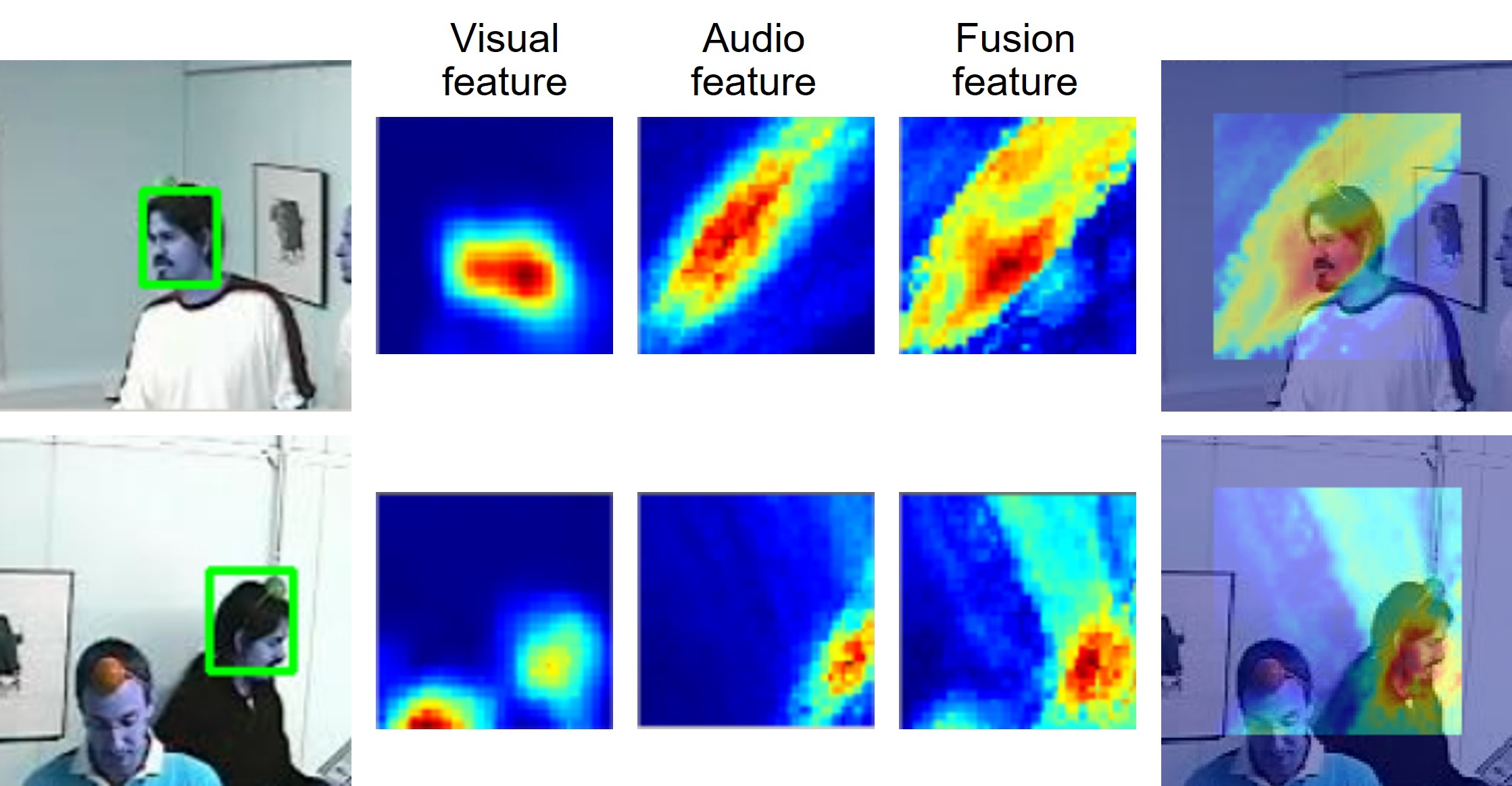} 
\caption{
Visualization of the visual feature, audio feature and fusion feature.
The green boxes represent tracking results for the specified speaker. The feature maps in the last column are resized and padded to the tracking area.
}
\label{fig-feature}
\vspace{-0.25cm}
\end{figure}
\subsection{Visualization Analysis}
The fusion feature map output by the cross-modal attention module is visualized as a heat map displayed in Figure~\ref{fig-fuisonmap}.
Highlighted attention focuses on key information in the search region, such as the area around the face.
As shown in $seq$24 on the right of Figure~\ref{fig-fuisonmap}, when disturbed by similar targets, part of the attention is shifted to the nearby speaker, but the primary attention is retained on the specified speaker.
Moreover, the visual and audio features before CMA and the fusion features after CMA are compared in Figure~\ref{fig-feature} to visualize the fusion effect.
In the example in the second row, the visual cues are influenced by a similar target, so attention is focused on the other speaker. 
Guided by reliable audio cues, the attention in the final fusion features shifts to the target being tracked.
Unlike simple feature addition, the proposed fusion method exploits the consistency and complementarity of multi-modalities to obtain more reliable fusion features.

\section{Conclusion}
In this study, we present a deep audio-visual fusion network for the challenging audio-visual speaker tracking task.
The proposed STNet relies on visual-guided acoustic measurement to map audio cues into a localization space consistent with the visual cues.
Heterogeneous cues are embedded into a unified feature space through visual CNN and audio CNN, and the interaction between different modalities is promoted through a cross-modal attention module.
The attention mechanism establishes cross-modal long-range correlations, allowing the tracker to adaptively focus on valuable information and derive more reliable fusion features.
Moreover, the STNet-based tracker is extended to more complex multi-speaker scenarios through a quality-aware module-based update/reset strategy.
The state-of-the-art performance on SOT and MOT demonstrates the effectiveness and robustness of the proposed tracker.
As an effective deep learning-based tracking framework, STNet verifies advantages of deep audio-visual fusion.
We look forward to providing more valuable insights into the field of audio-visual tracking.

\bibliographystyle{IEEEtran}
\footnotesize
\bibliography{mybib}

\vspace{-1cm}

\end{document}